\documentclass[10pt,twocolumn,letterpaper]{article}

\usepackage{iccv}
\usepackage{times}
\usepackage{epsfig}
\usepackage{graphicx}
\usepackage{amsmath}
\usepackage{amssymb}
\usepackage{capt-of,etoolbox}
\usepackage{siunitx}
\usepackage{textcomp} % Needed for apostrophes
\usepackage{authblk}
\usepackage[breaklinks=true,bookmarks=false]{hyperref}

\iccvfinalcopy % *** Uncomment this line for the final submission

\def\iccvPaperID{****} % *** Enter the ICCV Paper ID here

\begin{document}

%%%%%%%%% TITLE
\makeatletter
\renewcommand\AB@affilsepx{, \protect\Affilfont}
\makeatother
%%%%%%%%% TITLE
\title{Soft Prototyping Camera Designs for Car Detection Based on a Convolutional Neural Network}
\author[1,2]{Zhenyi Liu}
\author[1]{Trisha Lian}
\author[1]{Joyce Farrell}
\author[1]{Brian Wandell}

% \affil[1]{Stanford University, USA} 
% \affil[2]{Jilin University, China}
% \authorcr \{\tt zhenyiliu, tlian, jefarrel, bwandell\}@stanford.edu}
\affil[1]{Stanford University, USA} 
  \affil[2]{Jilin University, China \authorcr
  \{\tt zhenyiliu, tlian, jefarrel, wandell\}@stanford.edu}

\twocolumn[{ %
\renewcommand\twocolumn[1][]{#1}%
\maketitle
\begin{center}
     \centering
     \includegraphics[width=.95\textwidth]{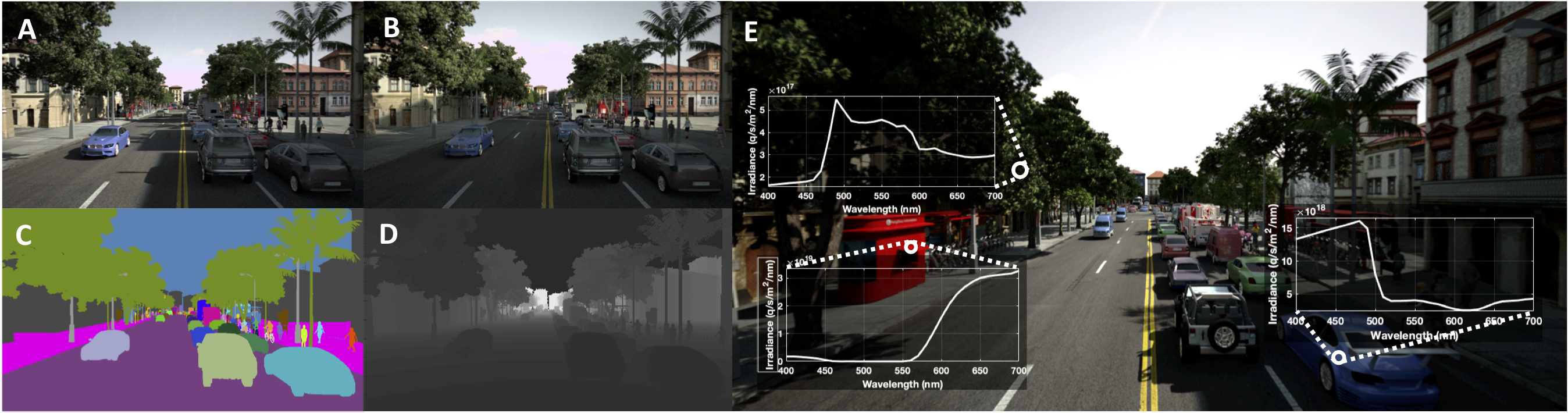}
\end{center}  %https://www.overleaf.com/project/5c93c425b089f6172ab153f3
     \centering
     \captionof{figure}{\textbf{Simulated scenes and metadata.} An ISE3d generated scene with sky maps measured in (A)  early afternoon and (B) late afternoon. The simulation creates pixel-level labeled images including (C) a panoptic segmentation, and (D) a depth map. (E) A simulated scene with superimposed graphs showing the spectral radiance at different image locations (circles). The spectral irradiance is used by the camera simulation to account for the wavelength effects of the lens and color filters.      
     \smallskip}
    %  \vspace{2mm}
\label{fig:figure1}
}]

%%%%%%%%% ABSTRACT
\begin{abstract}
\vspace{-6mm}
Imaging systems are increasingly used as input to convolutional neural networks (CNN) for object detection; we would like to design cameras that are optimized for this purpose. It is impractical to build different cameras and then acquire and label the necessary data for every potential camera design; creating software simulations of the camera in context (soft prototyping) is the only realistic approach. We implemented soft-prototyping tools that can quantitatively simulate image radiance and camera designs to create realistic images that are input to a convolutional neural network for car detection.  We used these methods to quantify the effect that critical hardware components (pixel size), sensor control (exposure algorithms) and image processing (gamma and demosaicing algorithms) have upon average precision of car detection. We quantify (a) the relationship between pixel size and the ability to detect cars at different distances, (b) the penalty for choosing a poor exposure duration, and (c) the ability of the CNN to perform car detection for a variety of post-acquisition processing algorithms. These results show that the optimal choices for car detection are not constrained by the same metrics used for image quality in consumer photography. It is better to evaluate camera designs for CNN applications using soft prototyping with task-specific metrics rather than consumer photography metrics.
\end{abstract}

%%%%%%%%% BODY TEXT
\vspace{-3mm}
\section{Introduction}

An imaging system should be evaluated with respect to a task. For example, we judge the image quality of digital cameras by how pleasing the final images appear to consumers.  Medical imaging systems are evaluated by how well they differentiate biological substrates and diagnose a disease.  One of the most critical aspects for automotive imaging systems is performance on object detection tasks.

The enormous diversity of biological eyes is strong evidence that different tasks are best served by different image system designs \cite{Land2001-md}. Even a single biological system, such as human vision, is comprised of multiple subsystems specialized for low and high light levels (rods, cones) or for high acuity tasks versus invoking attention (fovea, periphery) \cite{Wandell1995-cu}. Just as there is no optimal biological visual system for all tasks and conditions, no single camera design will be optimal.

Designing and building an imaging system for object detection is expensive, collecting image data is time-consuming, and annotating the images is labor-intensive. Hence, an empirical design-build-test loop is impractical for co-design of imaging systems and neural networks, and soft prototyping is required to reduce the time and expense. A prototyping system must combine quantitative computer graphics for creating accurate scene radiance with quantitative methods for simulating the imaging system. Such a system can simulate realistic camera images with accurate labels at each pixel (Figure \ref{fig:figure1}); these images can be used to train neural networks for object recognition and detection \cite{Wrenninge2018-dr,Liu2019-eo}. 

In this paper, we perform computational experiments to assess specific co-designs of cameras and a convolutional neural network for car detection. We first quantify how average precision measured for car detection varies with a key parameter of the camera hardware:  pixel size. Next, we analyze performance for variations in a critical sensor control algorithm:  exposure duration. Finally, we compare system performance for different choices of the image processing system: gamma correction and demosaicing.

%-------------------------------------------------------------------------
\section{Related work}

A number of groups have described the value in using computer graphics to create realistic images for machine-learning (ML). Two approaches have been used to generate training images: repurposing game engines \cite{Gaidon2016-wb,Johnson-Roberson2016-hz} and using physically based ray tracing methods \cite{Tsirikoglou2017-wo,Wrenninge2018-qp,Blasinski2018-nw,Liu2019-eo}. Game engines have the advantage of speed and the ability to efficiently produce video sequences.  The ray-tracing methods have the ability to produce quantitatively realistic scene spectral radiance that can be coordinated with camera simulations.  Given our objective, we use quantitative ray-tracing for our application.

Previous publications described an initial implementation of ISET3d, which includes procedural methods for generating complex automotive scenes \cite{Blasinski2018-nw,Liu2019-eo}. These scenes are rendered into images using a version of PBRT \cite{Pharr2016-fl} and ISETCam.  Those papers included preliminary evaluations of how camera design impacts CNN performance for pixel size and sensor type. We advance that work in several ways. The paper by \cite{Blasinski2018-nw} did not include procedural modeling and was restricted to relatively simple scenes; rendering was based on a version of PBRT \cite{Pharr2016-fl} that was subsequently improved with regards to material modeling. The paper by \cite{Liu2019-eo} used Faster RCNN that was pre-trained using camera data from BDD100k \cite{Yu2018-wr} and tested on ISET3D synthetic data. The work described here uses (a) a much larger and more complex collection of synthetic scenes, (b) network training with the appropriate synthetic data, (c) a new network, Mask R-CNN with a ResNet backbone, and (d) more extensive camera algorithm analyses.

%There is uncertainty about how well systems trained on synthetic data generalize to data obtained with cameras; similarly there are failures to generalize between data sets obtained using different cameras or even the same camera with different acquisition parameters \cite{Andreopoulos2012-ff,Torralba2011-pd}. 

There are alternative approaches to creating synthetic data. For example, it is possible to use a game engine and then apply domain adaptation methods to make the images more realistic \cite{Wu2004-ig,Duan2009-xv}. The realism is evaluated by assessing how well training on the synthetic dataset generalizes to a camera dataset. Domain randomization introduces random variations into the synthetic image with the hope that such perturbations force the network to focus on critical information \cite{Tremblay2018-uq}.  Domain stylization uses photorealistic image style transfer algorithms to transform synthetic images so that an independent network cannot discriminate synthetic and measured images \cite{Dundar2018-ta,Li2018-yf,Zhu2017-in}. Neither of these methods represent the scene spectral radiance, which is required to account for the impact of wavelength-dependent components, including the optics and sensors \cite{Blasinski2018-nw}. This severely limits the value of this approach for evaluating camera design.

% Buckler briefly.
To assess the impact of the image processing pipeline on CNN performance, \cite{Buckler2017-iy} used RGB images and an invertible model of the image processing pipeline to create a nominal sensor image.  This approach does not account for optics and sensor properties, such as pixel size or exposure algorithm.

% Carlson
Data augmentation can improve the generalization between synthetic and camera images. In this approach, a set of synthetic images are transformed by image processing operations that approximate camera effects, including blur, chromatic aberration, and color processing \cite{Carlson2018-nq,Carlson2019-oe}. The network becomes less sensitive to camera differences with augmented data. Data augmentation is complementary to our goal, which is to explore camera design to provide optimal network performance.

Finally, there is a standardization effort by the IEEE-P2020 to address attributes that contribute to image quality for automotive Advanced Driver Assistance Systems applications, as well as identifying existing metrics relating to these attributes \cite{Geese2018-xw,Zlokolica2018-gv}. Multiple metrics are under review, including consumer photography metrics such as signal-to-noise (SNR) and spatial frequency measures (MTF50). The soft prototyping environment we describe here clarifies limitations in using such metrics for assessing imaging systems designed for car detection by neural networks.

\section{Methods}
\begin{figure*}
\begin{center}
\includegraphics[width=0.9\linewidth]{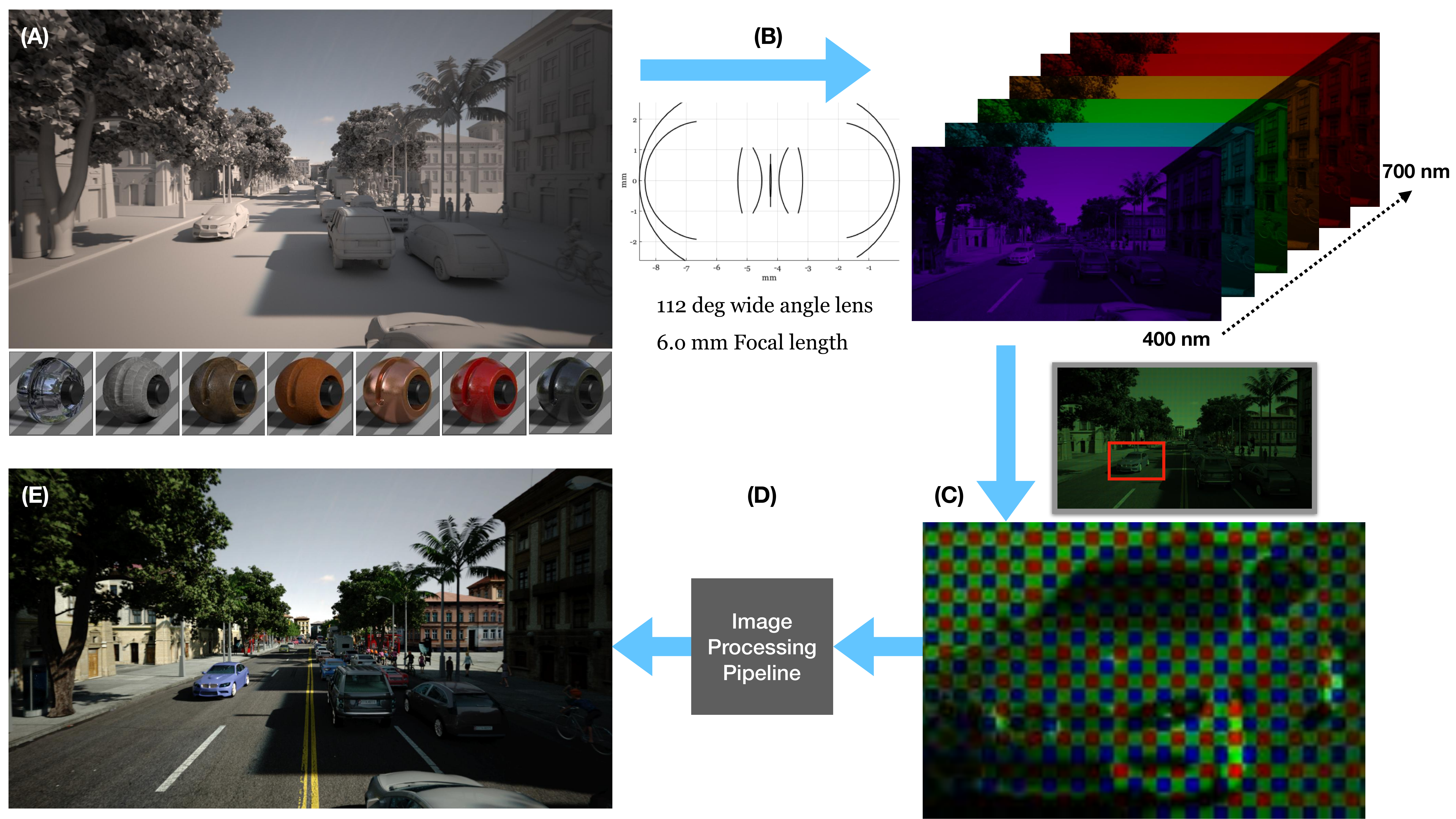}
\end{center}
  \caption{\textbf{The computational pipeline.}  (A) A three-dimensional scene, including objects and materials, is defined in the format used by Physically Based Ray Tracing (PBRT) software \cite{Pharr2016-fl}.  (B) The rays pass through a lens modeled as a series of surfaces with wavelength-dependent indices of refraction and a Monte Carlo based model of diffraction \cite{Freniere1999-ly,Smith2004-jk}. (C) Using the sensor specifications, ISETCam converts the spectral irradiance into an array of pixel voltages and then digital values \cite{Farrell2003-zi,Farrell2012-nl}. (D) Sensor data are converted into images by demosaicing and image processing. (E) The images with pixel level annotation are used to train and evaluate the car detection network.
}
\label{fig:figure2}
\end{figure*}

\subsection{Automotive scene simulation}
Simulated scene radiance data and sensor irradiance were generated for a collection of about 5000 city scenes using the ISET3d open-source software \cite{Liu2019-eo}.  The scenes were assembled stochastically from a database of nearly 100 car shapes, 3 bus shapes, 8 truck shapes, 80 pedestrians, 10 bicycles, more than 50 different static objects(trees, trash cans, traffic signs, billboards, etc.), four city blocks, suburban and more than 200 buildings.  The scenes are rendered with skymaps captured from 7:00 am to 7:00 pm at one minute intervals \cite{LeGendre2016-fh}. The source code for creating the images is available in the ISET3d package (\url{https://github.com/iset/iset3d}). The ray-tracing software is based on the Physically Based Ray Tracing \cite{Pharr2016-fl} and is available as a Docker container.

%-------------------------------------------------------------------------
\subsection{Camera simulation}
The ISETCam software converts the sensor spectral irradiance to RGB images (Figure \ref{fig:figure2}) \cite{Farrell2003-zi,Farrell2012-nl,Farrell2015-jb}. The sensor parameters for the simulations reported in this paper were based on the MT9V024 sensor manufactured by ON Semiconductor. This sensor is configured with 6-micron pixels, providing relatively high light sensitivity and signal-to-noise with a linear dynamic range of 55 dB.  The MT9V024 is designed for automotive machine vision applications with options for monochrome, RGB Bayer and RCCC color filter arrays. This sensor is used in commercial ADAS systems (e.g., MobilEye 630).

We modeled a range of pixel sizes, starting with a configuration that is typical for automotive sensors ($1280{\times}720$ pixels, RGB Bayer configuration). The sensor dye size is 2.1 x 3.85 mm.  We analyzed system performance over a range of pixel sizes and scene brightness levels.  For the pixel size experiments, we simulated a sensor with a fixed field-of-view (dye size). Consequently, the number of pixels varies inversely with the square of pixel size. The camera lens for these simulations was a wide angle (112 degree) multi-element design with 6 mm focal length.  The on-axis point spread of the lens has a full-width at half maximum of approximately \SI{1.5}{\micro\meter}, which was sufficient to support the smallest RGGB superpixel size (\SI{3.0}{\micro\meter}) in our simulations. 

\subsection{Soft prototyping validation}
Full system validation is not practical: this would require that we collect and label images with a simulated sensor and then synthesize scenes that match the collected images. It is possible, however, to validate quantitatively critical components of the soft-prototyping system, and we have done so. 

Sensor simulations were validated using a scene whose lights and surfaces were measured with a spectroradiometer and compared predicted data with data from a camera whose lens and sensor were modeled  \cite{Farrell2008-sc,Chen2009-bj}. The accuracy of the optics simulations were validated by comparing the simulated data with physical laws (diffraction-limited, Snell’s Law) and compared with Zemax calculations based on multi-element lens designs \cite{Lian2019-zt}. The geometry and materials models, in particular the light-surface interaction function captured in the bidirectional reflectance distribution function (BRDF), were derived from measurements of real materials. The 3D car models are derived from 3D scans or CAD designs of real cars. The spectral characteristics of the sky maps used to simulate daylight were measured using a multispectral lighting capture system and are also validated \cite{LeGendre2016-fh}. 

\subsection{Object Detection Network and Training}
For car detection on the camera, we chose Mask R-CNN \cite{Facebook_undated-wz} with ResNet50-FPN as the network backbone. Mask R-CNN is a state-of-the-art region-proposed convolutional neural network(R-CNN) designed to solve instance segmentation problems in computer vision \cite{He2017-bl}. It extends Faster R-CNN \cite{Ren2015-hy}, by adding a third branch that outputs the pixel level segmentation. Mask R-CNN also includes an ROIAlign layer to extract the feature map; the addition of this layer significantly improves detection accuracy. In this paper, we evaluate detection performance by measuring the overlap of the bounding box from Mask R-CNN with the bounding box of the labeled image data.

A pretrained model can be useful in some contexts, but to investigate the impact of  camera design it is best to train the network from scratch using the simulated camera data (e.g., pixel size, exposure algorithm, post-processing algorithm). Hence, the neural network only learns to interpret images from a specific camera under the specific imaging conditions. The sensor images were calculated from the spectral irradiance simulations using ISET3d and the ISETCam camera model. We divided the sensor images obtained from the scenes into three independent groups used for training, validation, and testing. We used 3000 sensor images for training, 700 held-out sensor images for validation at each checkpoint, and 750 held-out sensor images to measure system performance.

% overfitting issues addressed.
We trained the network for 60 epochs, saving the checkpoint every 5 epochs. We evaluated the network using the validation images at each of 12 checkpoints and we use the network parameters from the checkpoint with highest score. We then run this network on the test dataset at this checkpoint. We trained on 4 Nvidia P100 GPUs with batch size equal to 8 and evaluated on 1 GPU with batch size equal to 4. We started training with the learning rate set to 0.02, and we decreased the learning rate to 0.002 after 30 epochs. The model is trained and evaluated with only one class: car.

\subsection{Object detection metrics}
Car detection performance was assessed using the PASCAL AP@0.5IOU \cite{Everingham_undated-qm}, which we refer to as average precision (AP). When the network identifies a car within a bounding box, and that box overlaps with at least half of the area of a bounding box of the labeled pixels, we score the detection as correct (a hit), and otherwise the box is scored as an error (false alarm). The AP combines these two values and is equivalent to measuring the area under the receiver operating characteristic defined in classic signal detection theory \cite{Swets1978-vb}. 

In many cases we measure AP as a function of distance between the camera and the car. We can obtain this function because the soft-prototyping tools provide both the instances labels and a depth map of the scene for every pixel(Figure \ref{fig:figure1}).  We use the label (car) and depth (meters) to sort the cars in the test dataset.  We trace the curve by calculating the AP for all the cars within a 10 meter range. The distance (meters) at which the AP curve crosses 0.50 is a scalar summary value which we denote as the OD50.

\section{Experiments}
\subsection{Pixel size}
\begin{figure}[h]
   \begin{center}
     \includegraphics[width=1\linewidth]{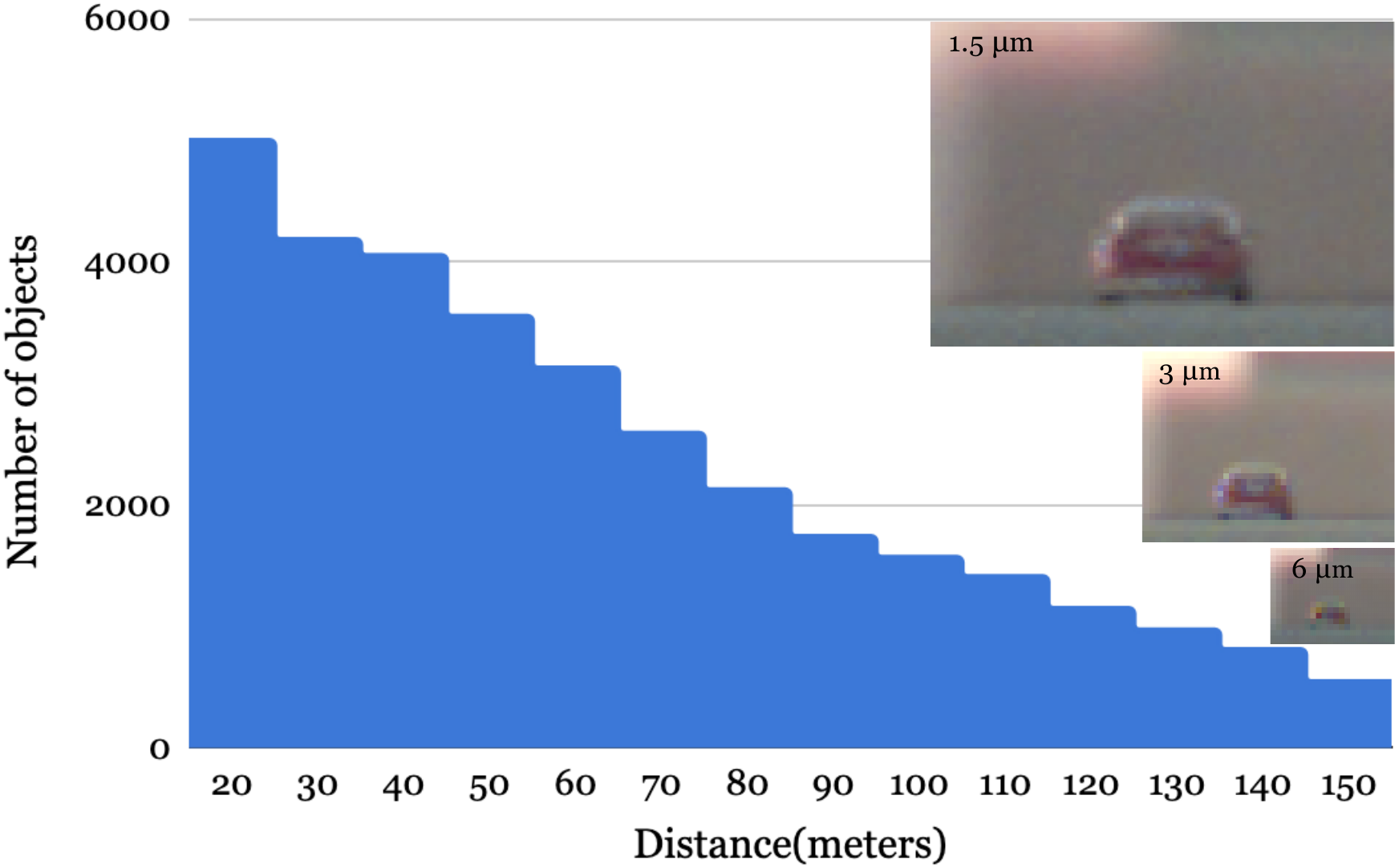}
   \end{center}
   \caption{\textbf{Number of cars for network training at different distances} The histogram represents the number of cars in the training set at each distance. The images at the right show simulations of a clear view of a car at \SI{150}{\meter} for each of the pixel sizes. The simulated pixel sizes span the range of typical modern day CMOS sensors; 3 and \SI{6}{\micro\meter} pixel sizes are common for current automotive sensors.}
  \label{fig:figure3}
\end{figure}

To assess how pixel size impacts the car detection AP, we simulated sensors with a range of pixel sizes that are typical for automotive applications: \SI{1.5}{\micro\meter} to \SI{6.0}{\micro\meter}.  The simulation labels the pixels arising from any car within \SI{150}{\meter}. The number of image pixels labeled as 'car' depends significantly on pixel size, occlusions, and distance to the car. There are roughly 33,000 labeled cars in the collection of training images. The number of labeled cars decreases with distance (Figure \ref{fig:figure3}, histogram), but is the same for all camera pixel sizes. As the pixel size increases, however, the number of labeled pixels per car decreases (Figure \ref{fig:figure3}, images). 

% Remake the car images so that the pixels in the image are the same. This means that 1.5um has 4x the number of pixels of 3 um and that has 4x the number of pixels of the 6 um.  Big 1.5 um on the top, then like an image pyramid underneath it.

\begin{figure}[h]
   \begin{center}
     \includegraphics[width=1\linewidth]{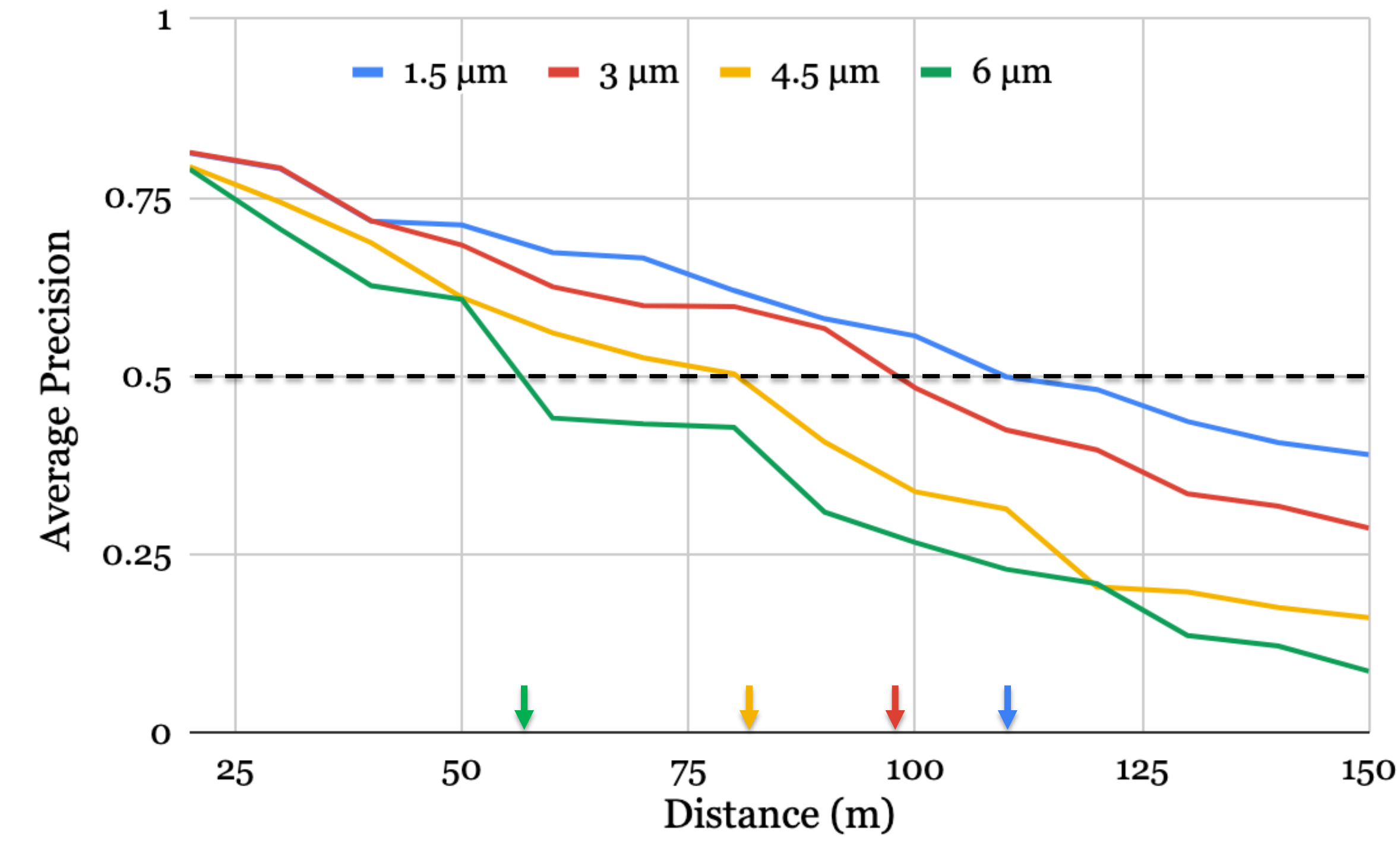}
   \end{center}
   \caption{\textbf{Car detection for cameras with different pixel sizes.} The solid curves shows the PASCAL AP@0.5IOU for car detection as a function of the distance between the camera and the car. Each curve simulates a different  pixel size. The corresponding colored arrows indicate the object distance for 50\% detection (OD50).}
  \label{fig:figure4}
\end{figure}

We trained the network separately using images from each pixel size, and we measured how AP declines with distance (Figure \ref{fig:figure4}). Detection performance using different pixel sizes declines gradually with distance. The rate of decline depends on pixel size. The distance where detection reaches $0.50$ increases non-linearly with pixel size, varying from \SI{55}{\meter} to \SI{110}{\meter}.

The gradual decline with distance is likely to be a consequence of the reduction in the number of pixels corresponding to each car: more distant cars occupy less of the field of view. This principle is similar to the Johnson criterion that is used to summarize device resolution \cite{Johnson1985-cs}. 

%\begin{table}[h]
%\begin{center}
%   \includegraphics[width=1\linewidth]{images/figure4.pdf}
%\end{center}
%   \caption{\textbf{Network generalization between cameras with different pixel sizes.} The AP for car detection at distances less than 150 m is shown for networks trained using data from cameras with different pixel sizes. The columns show the  pixel size used for training and the rows show the pixel size used for evaluation. Training on a small pixel size generalizes poorly to large pixel size data.  
%}
%\label{table:pscm}
% \end{table}

The choice of pixel size may be more significant for some applications than others.  For example, if the application is to survey cars within a distance between 1 and 40 meters (e.g., at a toll booth), pixel size may not be a significant factor. We make extensive use of the AP as a function of distance in this paper; it will be necessary to consider other task-specific metrics in future work. 

% The CNN parameters do not generalize well between data obtained with different cameras \cite{Carlson2018-nq}. We quantified the generalization by training the CNN with data from one pixel size and then testing on data with a different  pixel size (Figure \ref{fig:pscm}). The values are the average precision for car detection at all distances. In most cases, generalization between different pixels sizes is poor. The exception is training on 0.7 um generalizes well to 1.5 um. This is explained by the fact that these pixel sizes are equal to or smaller than the lens point spread. In all other cases, training at high resolution does not generalize to low resolution data, and conversely low resolution training does not generalize to high resolution data.  Networks can also fail to generalize between different empirical data sets, and this provides a strong motivation for using soft prototyping methods.

\subsection{Exposure algorithms}

The following experiments differ from the pixel size evaluations in two ways. First, we simulate a fixed \SI{3}{\micro\meter} pixel size. Second, we change the labeling policy to align with the conventional practice based on single camera data. The pixel size analysis used the knowledge from the simulation to label pixels: every car was labeled at every pixel size. Modern applications based on camera data typically label a pixel only if a human observer perceives the pixel as belonging to a car. We implemented this visibility constraint by labeling pixels only if they are part of a group of pixels with a bounding box of 10 x 15 pixels, big enough for a person to recognize. Including  perception as a criterion of the labeling policy removes \SI{24}{\%} of the car instances from the training and evaluation data but only \SI{0.14}{\%} of labeled pixels.  These are mainly distant or highly occluded cars. This restriction brings the evaluation into compliance with current practice and increases network performance (AP and OD50) by removing difficult-to-detect cars from the evaluation data.

Exposure value algorithms adjust the integration time and lens aperture to bring pixel responses into their operating range: pixel responses should be above the dark noise level and below the saturation level.  In this section, we analyze the impact of exposure-controlling algorithms on car detection, when the lens aperture is fixed.  We measure the impact using the average precision of CNN car detection in experiments with simulate scenes. The scenes are modeled at different times of the day and with a range of mean luminance levels, from extremely bright sunlight (\SI{500}{cd/m^2}) to very dark late afternoon (\SI{10}{cd/m^2}).  Under these conditions, the sensor illuminance of an f/\# 4 lens ranges from 10-500 lux. The diversity of surfaces and illumination in the simulated scenes generates images with dynamic ranges of 2-3 log units.

% End the distance axis at 150m in all the graphs.
\begin{figure}
\begin{center}
   \includegraphics[width=1\linewidth]{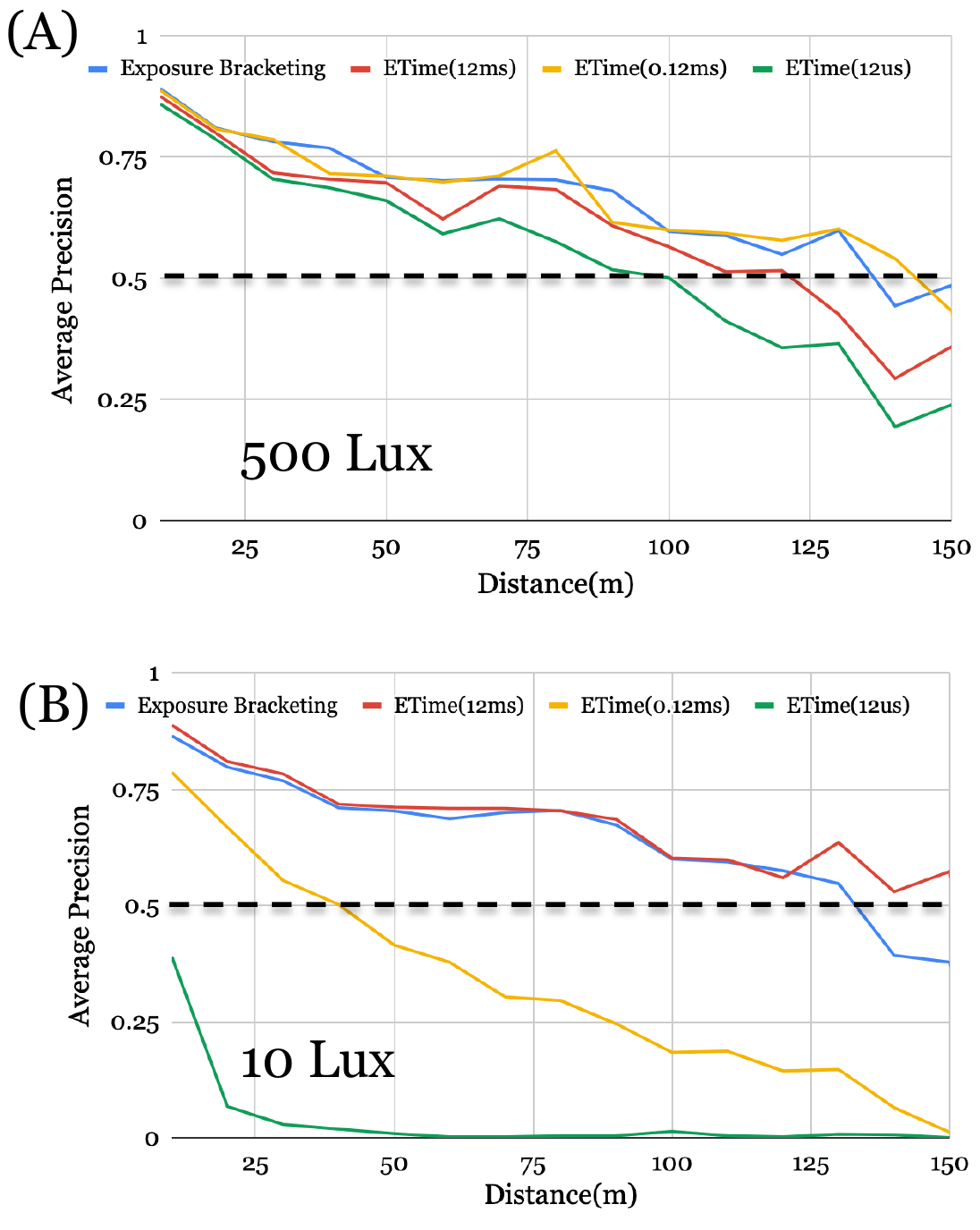}
\end{center}
   \caption{\textbf{Exposure duration has a significant impact on detection performance.} Average precision (AP) for car detection plotted as a function of the distance between the camera and the car with exposure time (\SI{12}{\ms}, \SI{0.12}{\ms}, \SI{12}{\micro\second}) as the parameter. AP data are calculated for networks trained on sensor data with mean illuminance of \SI{500}{\lux} (top) or with mean sensor illuminance of \SI{10}{\lux} (bottom).
}
\label{fig:figure5}
\end{figure}

\begin{figure}[h]
\begin{center}
   \includegraphics[width=1\linewidth]{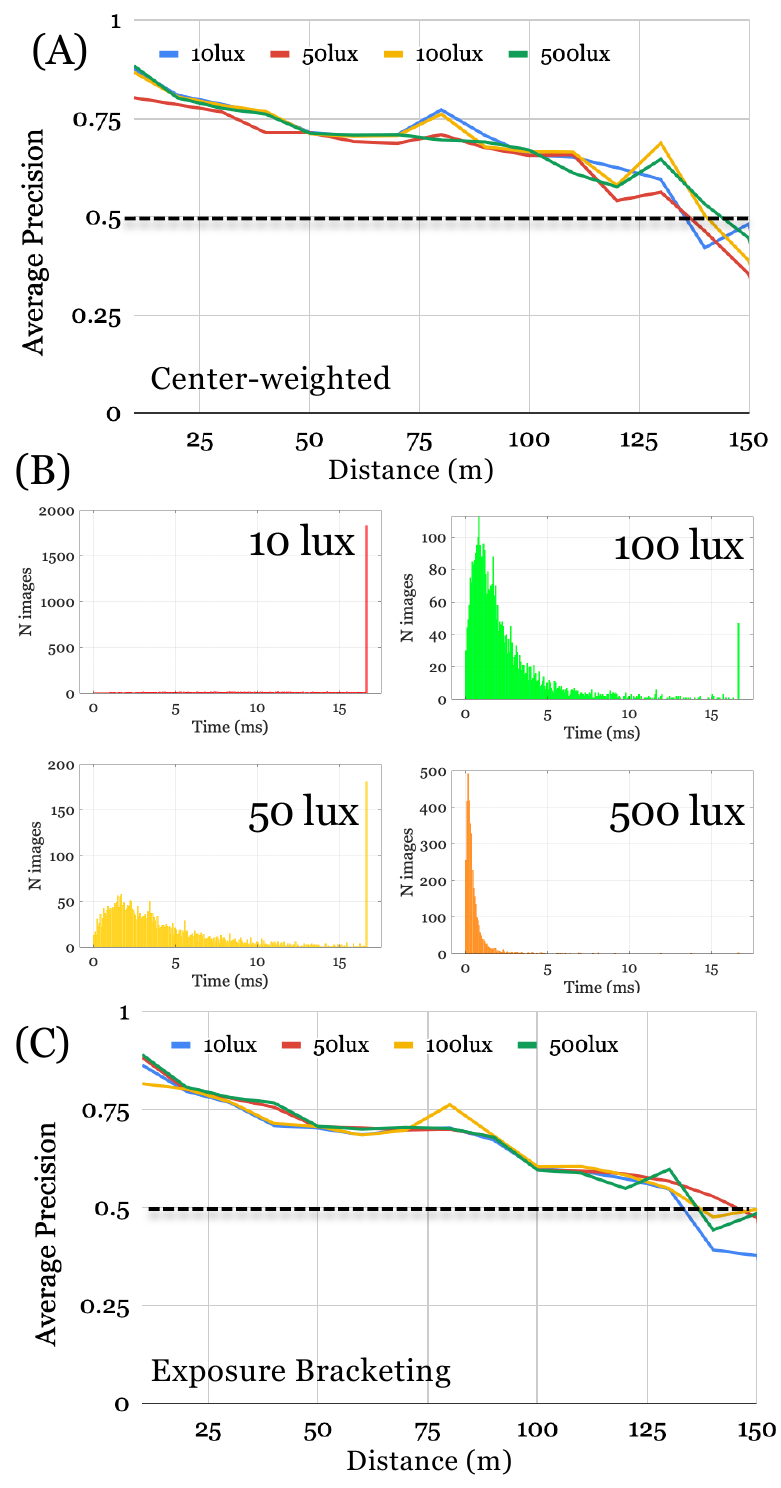}
\end{center}
   \centering\caption{\textbf{Comparison of center-weighted and exposure-bracketing exposure algorithms.} (A) Average precision for detecting cars in sensor images captured using a center-weighted auto-exposure algorithm, plotted as a function of the distance between the camera and the car for different sensor illuminance levels (B) Histogram of center-weighted exposure times for scenes with different sensor illuminance levels. (C) Average precision for detecting cars in sensor images captured using exposure-bracketing algorithms, plotted as a function of object distance for different sensor illuminance levels.
}
\label{fig:figure6}
\end{figure}

We quantified the performance of CNN models trained on images processed with two different exposure duration algorithms. The first is a center-weighted algorithm - the exposure duration is set so that a central region of the scene, about 100th of the whole image, falls within the pixel operating range. We impose a \SI{16}{\ms} upper bound based on an assumption that images are acquired at a rate of 60 frames per second (fps).  The second is an exposure-bracketing algorithm - we combine the data from three frames with different exposure times (\SI{12}{\ms}, \SI{0.12}{\ms}, \SI{12}{\micro\second}).  These images are combined by selecting the highest value prior to saturation, accounting for the exposure duration.  The exposure bracketing method with these durations extends the sensor dynamic range by two orders of magnitude.

Selecting an appropriate exposure duration has a significant impact on average precision; performance is significantly reduced when the algorithm chooses a poor exposure time (Figure \ref{fig:figure5}).  The curves in the two panels quantify the impact of choosing too long an exposure for a bright scene (Figure \ref{fig:figure5}A)  or too short an exposure in a dark scene (Figure \ref{fig:figure5}B). The penalty for choosing a poor exposure duration in the bright scene can be measured in terms of the OD50.  The best duration (\SI{0.12}{\ms}) has an OD50 of about 140 m, while the OD50 for the longer and shorter exposures are 125 m and 100 m.  Under the low light conditions, the best exposure duration (\SI{12}{\ms}) OD is about 130 m and choosing the wrong duration incurs an even larger penalty, with OD50 values of just 35 m and worse.

During informal experiments, we found that the region of interest of the exposure algorithm is important. Choosing a poor region or using the entire image led to poor results. The images used typically have a dynamic range of about 500:1, which fills up most of the pixel response range. Sub-optimal choice of the exposure duration puts part of the image beyond the pixel’s response range.  For example, if one uses the entire scene a short exposure time is often chosen to avoid saturation from the bright sky, and this choice reduces the visibility of cars within a shadowed portion of the image.  A center-weighted algorithm reduces the chance of an exposure duration error that impacts car detection.
\begin{figure*}[h]
\begin{center}
     \includegraphics[width=1\linewidth]{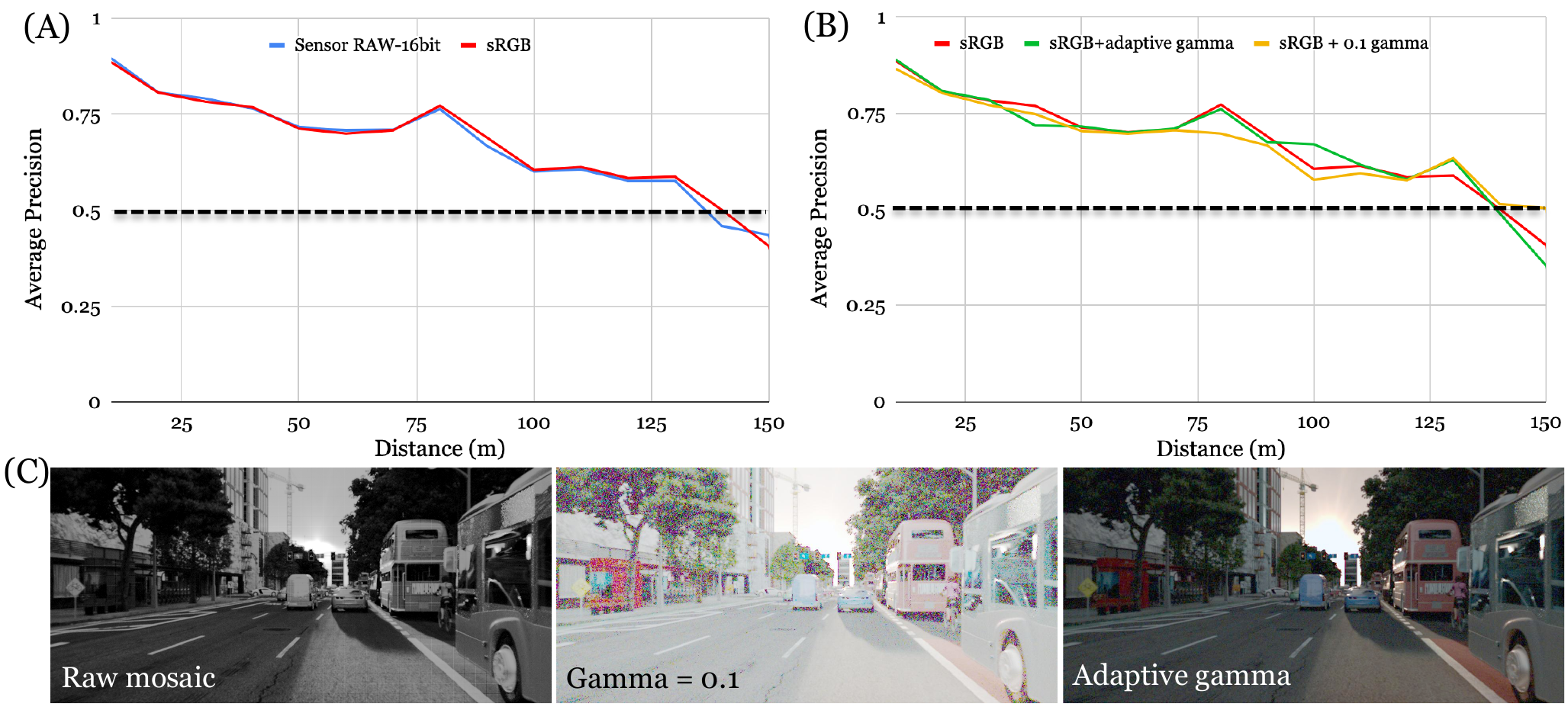}
\end{center}  
    \centering\caption{\textbf{The impact of image processing parameters on CNN performance.} (A) Average precision for detecting cars by networks trained on raw sensor data (blue) compared to average precision for networks trained on processed (sRGB) sensor data (red) plotted as a function of the distance between the camera and the car. (B) Average precision for networks trained on processed sensor data rendered with different gamma values plotted as a function of object distance. (C) Images of unprocessed sensor data (left image), and processed sensor data rendered with different gamma settings.
}
\label{fig:figure_gamma}
\end{figure*}
The exposure-bracketing algorithm reduces the chances of setting the wrong exposure duration: the algorithm acquires multiple captures at the cost of reading multiple frames and assembling the data into a single output at video rate. The three exposure durations used in the simulation were chosen to leave enough time for reading and assembling the output at \SI{60}{\hertz}. We compared the center-weighted algorithm and exposure bracketing for the collection of simulated images.  We plot car detection performance for scene illumination levels ranging from \SI{10}{\lux} to \SI{500}{\lux}  (Figure \ref{fig:figure6}).  The average performance on car detection does not differ substantially between the exposure-bracketing algorithm and the best exposure duration (\SI{0.12}{\ms} for panel A: yellow curve; and \SI{12}{\ms} for panel B: red curve).  The center-weighted exposure algorithm chooses many different exposure durations at each mean illumination level (see inset histograms).  

The center-weighted algorithm chooses an appropriate exposure time for the scenes in our collection with sensor illumination ranging between 10 and \SI{500}{\lux} (Figure \ref{fig:figure6}).  On average, these choices bring the image information about the cars into the pixel response range and car detection performance is not significantly different from the exposure-bracketing algorithm. While the averages agree, in certain cases using exposure-bracketing to increase the image dynamic range should have an advantage.  Below we examine one such edge case - for example system performance in specific cases such as a highly reflective (specular) surface in the middle of the road.

\subsection{Image processing}
Consumer photography imaging incorporates many post-acquisition processing algorithms to render sensor data. The algorithms include demosaicing, conversion of sensor data to a calibrated color space, illuminant correction algorithms, noise reduction methods, and mapping for dynamic range and color gamut.   The goal of these algorithms is to make an image that is pleasing to humans.  Preliminary studies have begun to assess whether designing systems for these image quality measures improves CNN performance. Because these algorithms take time and can be energy consuming, understanding their impact is likely to be an ongoing topic of investigation. To date there are no firm conclusions about their impact \cite{Buckler2017-iy,Blasinski2018-nw}. 

It is straightforward to implement image processing algorithms as part of the prototyping system, and we have quantified several post-acquisition processing steps with respect to car detection.  Figure \ref{fig:figure_gamma} compares the average precision of car detection in two cases.  In both cases we compare performance with a standard image processing pipeline (demosaicing, sensor conversion) that transforms the sensor data to JPEG format in sRGB space \cite{Nielsen1998-of}).  The first alternative eliminates the processing stream and the second alters the parameters of the processing.  

Neither alternative post-processing has a significant impact on the car detection performance in daytime. Results are very similar when the network is trained with an sRGB image or the unprocessed sensor mosaic data (center-weighted exposure algorithm, Figure \ref{fig:figure_gamma}A).  Maintaining the processing pipeline but using two different power functions (gamma mapping) to adjust the sRGB data also had no significant effect Figure \ref{fig:figure_gamma}B).  One method applied a fixed value (0.1) that brightened the high dynamic range images.  The second method chose gamma adaptively so that the mean value of the sensor data raised to a power (gamma) equals \SI{20}{\%} of the voltage swing.  The CNN training achieved the same performance level in each of these cases. This finding differs from \cite{Buckler2017-iy}, but they began with a network pre-trained on RGB images to evaluate the different image processing pipelines.
\begin{figure}[h]
  \begin{center}
     \includegraphics[width=0.9\linewidth]{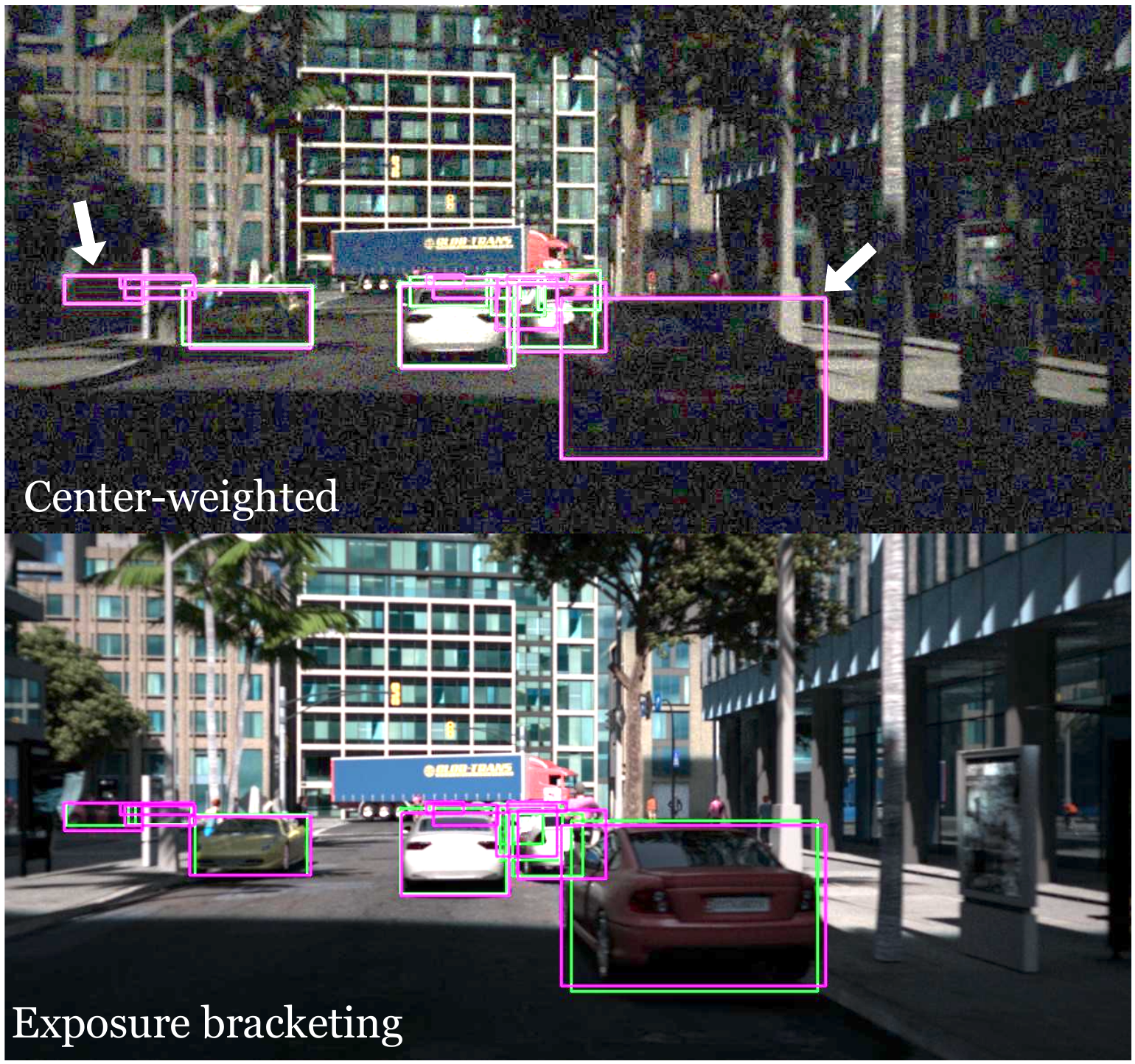}
  \end{center}
  \caption{\textbf{Examining individual cases.}  Images of the same scene captured with center-weighted (top) and exposure-bracketed (bottom) algorithms. Magenta bounding boxes indicate the locations of cars in the scene (ground truth) and green bounding boxes overlay the locations of cars that were detected.
}
  \label{fig:figure8}
\end{figure}
These post-acquisition transformations produce images that are extremely different to the human eye (Figure \ref{fig:figure_gamma}C) and yet they had no effect on the precision of CNN car detection. Measures of image quality are not a good guide for predicting image system performance for car detection.

\subsection{Edge cases}
For driving applications, the high cost of certain errors make relatively rare events very problematic. Hence, it is important to consider specific cases in addition to average precision. Soft prototyping tools enable us to isolate and analyze conditions that require special attention.

An important case that arises in our simulations is shown in Figure \ref{fig:figure8}. A white car in the center of the scene has a shiny (specular) surface that falls within the center-weighted region of interest. On the right side there is a darker (red) car within a shadow cast by a building.  The overlaid green bounding box indicates that the car is detected by the CNN when using the exposure-bracketing algorithm but not the center-weighted algorithm. The reasons is that the center-weighted algorithm selects a brief exposure duration and renders a noisy image in the shadowed region. The exposure-bracketing algorithm avoids this problem.  This situation does not arise often enough to reduce the average precision in our simulations, but such cases can be identified by searching for examples where the exposure-bracketing performance is better than center-weighted performance.

\section{Discussion and summary}
The design-build-test loop for automotive image systems is expensive and complex and this makes a soft prototyping tool necessary.  Additionally, it is typically impossible to know how variations in an isolated component will impact system performance. These requirements call for an end-to-end prototyping environment. Such a system can greatly speed innovation of image system architectures including choice of lens, sensors, and camera placement by measuring whole system performance on the relevant task. 

The open-source and freely available prototyping tools in ISET3d enables us to perform simulations that span scene definition, camera design, and network design. Our focus is on camera design and the experiments carried out in this paper explore aspects of the sensor hardware (pixel size), sensor control algorithms (exposure algorithms) and post-processing (demosaicing and gamma correction). Using the quantitative measure of average precision of car detection, and the summary measure of OD50, we quantified system performance. These experiments show that car detection algorithms identify cars at a distance approaching 115 meters for small (\SI{1.5}{\micro\meter}) pixels and 98 meters for larger (\SI{3}{\micro\meter}) pixels.  Further experiments with lens and sensor design might lead to improved performance. Center-weighted exposure produced good average precision, but exposure-bracketing algorithms have higher reliability and handle certain edge cases that are problematic for the single-exposure center-weighted acquisition.  Finally, we found no advantage for post-acquisition image processing for detecting cars in daytime conditions. When the network is trained from scratch on the relevant data, performance is nearly identical whether using digital values directly from the sensor mosaic or RGB values after post-acquisition processing. 

The consumer photography industry has developed many image quality tools, including SNR and MTF50. In principle these metrics might be used to assess image systems for car detection. We find, however, that there is a divergence between image quality metrics and CNN detection performance.  Images that would be intolerable in consumer photography - with extreme gamma values or without any post-acquisition processing - are effective inputs to CNN systems for car detection. End-to-end prototyping is preferable to using perceptual image quality metrics when predicting CNN performance. 

The diversity of biological systems inspires us to design and test increasingly specialized cameras. Soft prototyping can support experimentation with cameras optimized for motion, certain colors or shapes. It can also support experimentation using camera arrays with wide baselines that are specialized for depth estimation, or cameras that incorporate data about vehicle motion to identify still from moving objects and to use parallax to estimate depth. We hope that using end-to-end simulations in combination with task-specific performance metrics will yield insights that improve image system designs for this new generation of imaging systems.


\begin{thebibliography}{10}\itemsep=-1pt

\bibitem{Blasinski2018-nw}
H.~Blasinski, J.~Farrell, T.~Lian, Z.~Liu, and B.~Wandell.
\newblock Optimizing image acquisition systems for autonomous driving.
\newblock {\em Electronic Imaging}, 2018(5):161--1--161--7, Jan. 2018.

\bibitem{Buckler2017-iy}
M.~Buckler, S.~Jayasuriya, and A.~Sampson.
\newblock Reconfiguring the imaging pipeline for computer vision, 2017.

\bibitem{Carlson2018-nq}
A.~Carlson, K.~A. Skinner, and M.~Johnson-Roberson.
\newblock Modeling camera effects to improve deep vision for real and synthetic
  data.
\newblock {\em arXiv preprint arXiv:1803. 07721}, 2018.

\bibitem{Carlson2019-oe}
A.~Carlson, K.~A. Skinner, R.~Vasudevan, and M.~Johnson-Roberson.
\newblock Sensor transfer: Learning optimal sensor effect image augmentation
  for {Sim-to-Real} domain adaptation.
\newblock {\em IEEE Robotics and Automation Letters}, pages 1--1, 2019.

\bibitem{Chen2009-bj}
J.~Chen, K.~Venkataraman, D.~Bakin, B.~Rodricks, R.~Gravelle, P.~Rao, and
  Y.~Ni.
\newblock Digital camera imaging system simulation.
\newblock {\em IEEE Trans. Electron Devices}, 56(11):2496--2505, Nov. 2009.

\bibitem{Duan2009-xv}
L.~Duan, I.~W. Tsang, D.~Xu, and T.-S. Chua.
\newblock Domain adaptation from multiple sources via auxiliary classifiers.
\newblock In {\em Proceedings of the 26th Annual International Conference on
  Machine Learning}, ICML '09, pages 289--296, New York, NY, USA, 2009. ACM.

\bibitem{Dundar2018-ta}
A.~Dundar, M.-Y. Liu, T.-C. Wang, J.~Zedlewski, and J.~Kautz.
\newblock Domain stylization: A strong, simple baseline for synthetic to real
  image domain adaptation.
\newblock {\em arXiv [cs.CV]}, July 2018.

\bibitem{Everingham_undated-qm}
M.~Everingham, L.~van Gool, C.~Williams, J.~Winn, and Z.~A.
\newblock Pascal visual object classes.
\newblock \url{http://host.robots.ox.ac.uk/pascal/VOC/}.
\newblock Accessed: 2019-NA-NA.

\bibitem{Facebook_undated-wz}
{Facebook}.
\newblock maskrcnn-benchmark.

\bibitem{Farrell2008-sc}
J.~Farrell, M.~Okincha, and M.~Parmar.
\newblock Sensor calibration and simulation.
\newblock In {\em Digital Photography {IV}}, volume 6817, page 68170R.
  International Society for Optics and Photonics, Mar. 2008.

\bibitem{Farrell2012-nl}
J.~E. Farrell, P.~B. Catrysse, and B.~A. Wandell.
\newblock Digital camera simulation.
\newblock {\em Appl. Opt.}, 51(4):A80--90, Feb. 2012.

\bibitem{Farrell2015-jb}
J.~E. Farrell and B.~A. Wandell.
\newblock Image systems simulation.
\newblock {\em Handbook of Digital Imaging}, 1:373--400, 2015.

\bibitem{Farrell2003-zi}
J.~E. Farrell, F.~Xiao, P.~B. Catrysse, and B.~A. Wandell.
\newblock A simulation tool for evaluating digital camera image quality.
\newblock In {\em Image Quality and System Performance}, volume 5294, pages
  124--131. International Society for Optics and Photonics, Dec. 2003.

\bibitem{Freniere1999-ly}
E.~R. Freniere, G.~Groot~Gregory, and R.~A. Hassler.
\newblock Edge diffraction in monte carlo ray tracing.
\newblock In {\em Optical Design and Analysis Software}, volume 3780, pages
  151--158. International Society for Optics and Photonics, Sept. 1999.

\bibitem{Gaidon2016-wb}
A.~Gaidon, Q.~Wang, Y.~Cabon, and E.~Vig.
\newblock Virtual worlds as proxy for multi-object tracking analysis.
\newblock In {\em Proceedings of the {IEEE} conference on computer vision and
  pattern recognition}, pages 4340--4349, 2016.

\bibitem{Geese2018-xw}
M.~Geese, U.~Seger, and A.~Paolillo.
\newblock Detection probabilities: Performance prediction for sensors of
  autonomous vehicles.
\newblock {\em Electronic Imaging}, 2018.

\bibitem{He2017-bl}
K.~He, G.~Gkioxari, P.~Doll{\'a}r, and R.~Girshick.
\newblock Mask {R-CNN}.
\newblock {\em arXiv [cs.CV]}, Mar. 2017.

\bibitem{Johnson1985-cs}
J.~Johnson.
\newblock Analysis of image forming systems.
\newblock In W.~L.~W. R.~Barry~Johnson, editor, {\em Selected Papers on
  Infrared Design, Part One and Part Two.}, volume 513 of {\em Selected papers
  on infrared design. Part I and II}, page 761. Publisher, SPIE-The
  International Society for Optical Engineering, Bellingham, Washington, 1985.

\bibitem{Johnson-Roberson2016-hz}
M.~Johnson-Roberson, C.~Barto, R.~Mehta, S.~N. Sridhar, K.~Rosaen, and
  R.~Vasudevan.
\newblock Driving in the matrix: Can virtual worlds replace {Human-Generated}
  annotations for real world tasks?
\newblock Oct. 2016.

\bibitem{Land2001-md}
M.~F. Land.
\newblock {THE} {OPTICS} {OF} {ANIMAL} {EYES}, 2001.

\bibitem{LeGendre2016-fh}
C.~LeGendre, X.~Yu, D.~Liu, J.~Busch, A.~Jones, S.~Pattanaik, and P.~Debevec.
\newblock Practical multispectral lighting reproduction.
\newblock {\em ACM Trans. Graph.}, 35(4):32, July 2016.

\bibitem{Li2018-yf}
Y.~Li, M.-Y. Liu, X.~Li, M.-H. Yang, and J.~Kautz.
\newblock A {Closed-Form} solution to photorealistic image stylization: 15th
  european conference, munich, germany, september 8--14, 2018, proceedings,
  part {III}.
\newblock In V.~Ferrari, M.~Hebert, C.~Sminchisescu, and Y.~Weiss, editors,
  {\em Computer Vision -- {ECCV} 2018}, volume 11207 of {\em Lecture Notes in
  Computer Science}, pages 468--483. Springer International Publishing, Cham,
  2018.

\bibitem{Lian2019-zt}
T.~Lian, K.~J. MacKenzie, D.~H. Brainard, N.~P. Cottaris, and B.~A. Wandell.
\newblock Ray tracing {3D} spectral scenes through human optics models.
\newblock Mar. 2019.

\bibitem{Liu2019-eo}
Z.~Liu, M.~Shen, J.~Zhang, S.~Liu, H.~Blasinski, T.~Lian, and B.~Wandell.
\newblock A system for generating complex physically accurate sensor images for
  automotive applications.
\newblock Feb. 2019.

\bibitem{Nielsen1998-of}
M.~Nielsen and M.~Stokes.
\newblock The creation of the {sRGB} {ICC} profile.
\newblock {\em Color and Imaging Conference}, 1998(1):253--257, 1998.

\bibitem{Pharr2016-fl}
M.~Pharr, W.~Jakob, and G.~Humphreys.
\newblock {\em Physically Based Rendering: From Theory to Implementation}.
\newblock Morgan Kaufmann, Sept. 2016.

\bibitem{Ren2015-hy}
S.~Ren, K.~He, R.~Girshick, and J.~Sun.
\newblock Faster {R-CNN}: Towards {Real-Time} object detection with region
  proposal networks.
\newblock In C.~Cortes, N.~D. Lawrence, D.~D. Lee, M.~Sugiyama, and R.~Garnett,
  editors, {\em Advances in Neural Information Processing Systems 28}, pages
  91--99. Curran Associates, Inc., 2015.

\bibitem{Smith2004-jk}
W.~Smith.
\newblock {\em Modern Lens Design}.
\newblock McGraw Hill Professional, Oct. 2004.

\bibitem{Swets1978-vb}
J.~A. Swets and D.~M. Green.
\newblock Applications of signal detection theory.
\newblock In H.~L. Pick, H.~W. Leibowitz, J.~E. Singer, A.~Steinschneider, and
  H.~W. Stevenson, editors, {\em Psychology: From Research to Practice}, pages
  311--331. Springer US, Boston, MA, 1978.

\bibitem{Tremblay2018-uq}
J.~Tremblay, A.~Prakash, D.~Acuna, M.~Brophy, V.~Jampani, C.~Anil, T.~To,
  E.~Cameracci, S.~Boochoon, and S.~Birchfield.
\newblock Training deep networks with synthetic data: Bridging the reality gap
  by domain randomization.
\newblock In {\em 2018 {IEEE/CVF} Conference on Computer Vision and Pattern
  Recognition Workshops ({CVPRW})}, pages 1082--10828, June 2018.

\bibitem{Tsirikoglou2017-wo}
A.~Tsirikoglou, J.~Kronander, M.~Wrenninge, and J.~Unger.
\newblock Procedural modeling and physically based rendering for synthetic data
  generation in automotive applications.
\newblock Oct. 2017.

\bibitem{Wandell1995-cu}
B.~A. Wandell.
\newblock {\em Foundations of vision}.
\newblock Sinauer Associates, Sunderland, MA, 1995.

\bibitem{Wrenninge2018-dr}
M.~Wrenninge and J.~Unger.
\newblock Synscapes: A photorealistic synthetic dataset for street scene
  parsing.
\newblock Oct. 2018.

\bibitem{Wrenninge2018-qp}
M.~Wrenninge and J.~Unger.
\newblock Synscapes: A photorealistic synthetic dataset for street scene
  parsing.
\newblock Oct. 2018.

\bibitem{Wu2004-ig}
P.~Wu and T.~G. Dietterich.
\newblock Improving {SVM} accuracy by training on auxiliary data sources.
\newblock In {\em Proceedings of the Twenty-first International Conference on
  Machine Learning}, ICML '04, pages 110--, New York, NY, USA, 2004. ACM.

\bibitem{Yu2018-wr}
F.~Yu, W.~Xian, Y.~Chen, F.~Liu, M.~Liao, V.~Madhavan, and T.~Darrell.
\newblock {{BDD100K}}: A diverse driving video database with scalable
  annotation tooling.
\newblock {\em arXiv [cs.CV]}, May 2018.

\bibitem{Zhu2017-in}
J.~Zhu, T.~Park, P.~Isola, and A.~A. Efros.
\newblock Unpaired {Image-to-Image} translation using {Cycle-Consistent}
  adversarial networks.
\newblock In {\em 2017 {IEEE} International Conference on Computer Vision
  ({ICCV})}, pages 2242--2251, Oct. 2017.

\bibitem{Zlokolica2018-gv}
V.~Zlokolica, M.~Griffin, A.~Casey, D.~Solera, B.~Deegan, P.~Denny, and
  B.~Dever.
\newblock Visual quality evaluation of the multi-camera visualization in
  automotive surround view systems.
\newblock {\em Electronic Imaging}, 2018(17):1--6, 2018.

\end{thebibliography}


{\small


}
\end{document}